\documentclass[aos,preprint]{imsart}
\pdfoutput=1

\usepackage[OT1]{fontenc}
\usepackage[round]{natbib}
\usepackage[colorlinks,citecolor=blue,urlcolor=blue]{hyperref}
\usepackage{hypernat}
\usepackage{amsmath}
\usepackage{amssymb}
\usepackage{amsthm}
\usepackage{subfig}
\usepackage{graphicx}

\startlocaldefs
\usepackage{macros}
\endlocaldefs

\setattribute{journal}{name}{}

\usepackage[
  top=2.5cm,
  bottom=2.5cm,
  left=3cm,
  right=3cm
]{geometry}

\begin{document}
\begin{frontmatter}
  \title{Practical Bayesian Optimization of Machine Learning Algorithms} 
  \runtitle{Practical Bayesian Optimization of ML Algorithms}%

  \begin{aug}

    \author{\fnms{Jasper} \snm{Snoek}%
      \ead[label=e1]{jasper@cs.toronto.edu}},%
    \address{Jasper Snoek\\
      Department of Computer Science\\
      University of Toronto\\
      \printead{e1}}
    \author{\fnms{Hugo} \snm{Larochelle}%
      \ead[label=e2]{hugo.larochelle@usherbrooke.ca}}%
    \address{Hugo Larochelle\\
      D\'{e}partement d'informatique\\
      Universit\'{e} de Sherbrooke\\
      \printead{e2}}
    \and
    \author{\fnms{Ryan P.} \snm{Adams}%
      \ead[label=e3]{rpa@seas.harvard.edu}}%
    \address{Ryan P. Adams\\
      School of Engineering and Applied Sciences\\
      Harvard University\\
      \printead{e3}}

    \affiliation{University of Toronto,
      Universit\'{e} de Sherbrooke
      and Harvard University}

    \runauthor{Snoek, Larochelle and Adams}
  \end{aug}

\begin{abstract}
  Machine learning algorithms frequently require careful tuning of
  model hyperparameters, regularization terms, and optimization
  parameters.  Unfortunately, this tuning is often a ``black art''
  that requires expert experience, unwritten rules of thumb, or
  sometimes brute-force search. Much more appealing is the idea of
  developing automatic approaches which can optimize the performance
  of a given learning algorithm to the task at hand.  In this work, we
  consider the automatic tuning problem within the framework of
  Bayesian optimization, in which a learning algorithm's
  generalization performance is modeled as a sample from a Gaussian
  process (GP).  The tractable posterior distribution induced by the
  GP leads to efficient use of the information gathered by previous
  experiments, enabling optimal choices about what parameters to try
  next.  Here we show how the effects of the Gaussian process prior
  and the associated inference procedure can have a large impact on
  the success or failure of Bayesian optimization.  We show that
  thoughtful choices can lead to results that exceed expert-level
  performance in tuning machine learning algorithms.  We also describe
  new algorithms that take into account the variable cost (duration)
  of learning experiments and that can leverage the presence of
  multiple cores for parallel experimentation. We show that these
  proposed algorithms improve on previous automatic procedures and can
  reach or \emph{surpass} human expert-level optimization on a diverse
  set of contemporary algorithms including latent Dirichlet
  allocation, structured SVMs and convolutional neural networks.
\end{abstract}
\end{frontmatter}

\section{Introduction}
Machine learning algorithms are rarely parameter-free; whether via the
properties of a regularizer, the hyperprior of a generative model, or
the step size of a gradient-based optimization, learning procedures
almost always require a set of high-level choices that significantly
impact generalization performance.  As a practitioner, one is usually
able to specify the general framework of an inductive bias much more
easily than the particular weighting that it should have relative to
training data.  As a result, these high-level parameters are often
considered a nuisance, making it desirable to develop algorithms with
as few of these ``knobs'' as possible.

Another, more flexible take on this issue is to view the optimization
of high-level parameters as a procedure to be automated. Specifically,
we could view such tuning as the optimization of an unknown black-box
function that reflects generalization performance and invoke
algorithms developed for such problems.  These optimization problems
have a somewhat different flavor than the low-level objectives one
often encounters as part of a training procedure: here function
evaluations are very expensive, as they involve running the primary
machine learning algorithm to completion.  In this setting where
function evaluations are expensive, it is desirable to spend
computational time making better choices about where to seek the best
parameters. Bayesian optimization~\citep{Mockus1978} provides an
elegant approach and has been shown to outperform other state of the
art global optimization algorithms on a number of challenging
optimization benchmark functions \citep{Jones2001}.  For continuous
functions, Bayesian optimization typically works by assuming the
unknown function was sampled from a Gaussian process (GP) and
maintains a posterior distribution for this function as observations
are made.  In our case, these observations are the measure of
generalization performance under different settings of the
hyperparameters we wish to optimize. To pick the hyperparameters of
the next experiment, one can optimize the expected improvement
(EI)~\citep{Mockus1978} over the current best result or the Gaussian
process upper confidence bound (UCB)\citep{Srinivas2010}. EI and UCB
have been shown to be efficient in the number of function evaluations
required to find the global optimum of many multimodal black-box
functions~\citep{Bull2011, Srinivas2010}.

Machine learning algorithms, however, have certain characteristics
that distinguish them from other black-box optimization problems.
First, each function evaluation can require a variable amount of time:
training a small neural network with 10 hidden units will take less
time than a bigger network with 1000 hidden units.  Even without
considering duration, the advent of cloud computing makes it possible
to quantify economically the cost of requiring large-memory machines
for learning, changing the actual cost in dollars of an experiment
with a different number of hidden units.  It is desirable to
understand how to include a concept of cost into the optimization
procedure.  Second, machine learning experiments are often run in
parallel, on multiple cores or machines.  We would like to build
Bayesian optimization procedures that can take advantage of this
parallelism to reach better solutions more quickly.

In this work, our first contribution is the identification of good
practices for Bayesian optimization of machine learning algorithms.
In particular, we argue that a fully Bayesian treatment of the GP
kernel parameters is of critical importance to robust results, in
contrast to the more standard procedure of optimizing hyperparameters
(e.g.\ \citet{BergstraJ2011}).  We also examine the impact of the
kernel itself and examine whether the default choice of the
squared-exponential covariance function is appropriate.  Our second
contribution is the description of a new algorithm that accounts for
cost in experiments.  Finally, we also propose an algorithm that can
take advantage of multiple cores to run machine learning experiments
in parallel.

\section{Bayesian Optimization with Gaussian Process Priors}
As in other kinds of optimization, in Bayesian optimization we are
interested in finding the minimum of a function~$f(\brmx)$ on some
bounded set~$\mcX$, which we will take to be a subset of~$\reals^D$.
What makes Bayesian optimization different from other procedures is
that it constructs a probabilistic model for~$f(\brmx)$ and then
exploits this model to make decisions about where in~$\mcX$ to next
evaluate the function, while integrating out uncertainty.  The
essential philosophy is to use \emph{all} of the information available
from previous evaluations of~$f(\brmx)$ and not simply rely on local
gradient and Hessian approximations.  This results in a procedure that
can find the minimum of difficult non-convex functions with relatively
few evaluations, at the cost of performing more computation to
determine the next point to try.  When evaluations of~$f(\brmx)$ are
expensive to perform --- as is the case when it requires training a
machine learning algorithm --- it is easy to justify some extra
computation to make better decisions.  For an overview of the Bayesian
optimization formalism, see, e.g., \citet{brochu-etal-2010a}.  In this
section we briefly review the general Bayesian optimization approach,
before discussing our novel contributions in
Section~\ref{sec:practical}.

There are two major choices that must be made when performing Bayesian
optimization.  First, one must select a prior over functions that will
express assumptions about the function being optimized.  For this we
choose the Gaussian process prior, due to its flexibility and
tractability.  Second, we must choose an \emph{acquisition function},
which is used to construct a utility function from the model
posterior, allowing us to determine the next point to evaluate.

\subsection{Gaussian Processes}
The Gaussian process (GP) is a convenient and powerful prior
distribution on functions, which we will take here to be of the
form~${f:\mcX\to\reals}$.  The GP is defined by the property that any
finite set of~$N$ points~$\{\brmx_n\in\mcX\}^N_{n=1}$ induces a
multivariate Gaussian distribution on~$\reals^N$.  The~$n$th of these
points is taken to be the function value~$f(\brmx_n)$, and the elegant
marginalization properties of the Gaussian distribution allow us to
compute marginals and conditionals in closed form.  The support and
properties of the resulting distribution on functions are determined
by a mean function~${m:\mcX\to\reals}$ and a positive definite
covariance function~${K:\mcX\times\mcX\to\reals}$.  We will discuss
the impact of covariance functions in Section~\ref{sec:covariances}.
For an overview of Gaussian processes, see~\citet{Rasmussen2006}.

\subsection{Acquisition Functions for Bayesian Optimization}
We assume that the function~$f(\brmx)$ is drawn from a Gaussian
process prior and that our observations are of the
form~$\{\brmx_n,y_n\}^N_{n=1}$,
where~${y_n\sim\distNorm(f(\brmx_n),\nu)}$ and~$\nu$ is the variance
of noise introduced into the function observations.  This prior and
these data induce a posterior over functions; the acquisition
function, which we denote by~${a:\mcX\to\reals^{+}}$, determines what
point in~$\mcX$ should be evaluated next via a proxy
optimization~${\brmx_{\sf{next}} = \argmax_{\brmx} a(\brmx)}$, where
several different functions have been proposed.  In general, these
acquisition functions depend on the previous observations,
as well as the GP hyperparameters; we denote this dependence
as~$a(\brmx\aqhist)$.  There are several popular
choices of acquisition function.  Under the Gaussian process prior,
these functions depend on the model solely through its predictive mean
function~$\mu(\brmx\aqhist)$ and predictive
variance function~$\sigma^2(\brmx\aqhist)$.  In
the proceeding, we will denote the best current value
as~${\brmx_{\sf{best}}=\argmin_{\brmx_n}f(\brmx_n)}$,~$\Phi(\cdot)$
will denote the cumulative distribution function of the standard
normal, and~$\phi(\cdot)$ will denote the standard normal density
function.

\paragraph{Probability of Improvement}
One intuitive strategy is to maximize the probability of improving
over the best current value \citep{kushner-1964a}.  Under the GP this
can be computed analytically as
\begin{align}
  a_{\sf{PI}}(\brmx\aqhist) &= \Phi(\gamma(\brmx))
  &
  \gamma(\brmx) &= \frac{f(\brmx_{\sf{best}}) -\mu(\brmx\aqhist)}{\sigma(\brmx\aqhist)}.
\end{align}

\paragraph{Expected Improvement}
Alternatively, one could choose to maximize the expected improvement (EI)
over the current best.  This also has closed form under the Gaussian
process:
\begin{align}
  a_{\sf{EI}}(\brmx\aqhist) = \sigma(\brmx\aqhist)
  \left(
  \gamma(\brmx)\,\Phi(\gamma(\brmx)) + \distNorm(\gamma(\brmx)\,;\,0,1)
  \right)
\end{align}

\paragraph{GP Upper Confidence Bound}
A more recent development is the idea of exploiting lower confidence
bounds (upper, when considering maximization) to construct acquisition
functions that minimize regret over the course of their optimization
\citep{Srinivas2010}.  These acquisition functions have the form
\begin{align}
  a_{\sf{LCB}}(\brmx\aqhist) &= \mu(\brmx\aqhist) - \kappa\,\sigma(\brmx\aqhist),
\end{align}
with a tunable~$\kappa$ to balance exploitation against exploration.

In this work we will focus on the expected improvement criterion, as
it has been shown to be better-behaved than probability of
improvement, but unlike the method of GP upper confidence bounds
(GP-UCB), it does not require its own tuning parameter.  We have found
expected improvement to perform well in minimization problems, but
wish to note that the regret formalization is more appropriate for
many settings.  We perform a direct comparison between our EI-based
approach and GP-UCB in Section~\ref{sec:branin}.

\section{Practical Considerations for Bayesian Optimization of Hyperparameters}

\label{sec:practical}
Although an elegant framework for optimizing expensive functions,
there are several limitations that have prevented it from becoming a
widely-used technique for optimizing hyperparameters in machine
learning problems.  First, it is unclear for practical problems what
an appropriate choice is for the covariance function and its
associated hyperparameters.  Second, as the function evaluation itself
may involve a time-consuming optimization procedure, problems may vary
significantly in duration and this should be taken into account.
Third, optimization algorithms should take advantage of multi-core
parallelism in order to map well onto modern computational
environments.  In this section, we propose solutions to each of these
issues.

\subsection{Covariance Functions and Treatment of Covariance Hyperparameters}
\label{sec:covariances}
The power of the Gaussian process to express a rich distribution on
functions rests solely on the shoulders of the covariance function.
While non-degenerate covariance functions correspond to infinite
bases, they nevertheless can correspond to strong assumptions
regarding likely functions.  In particular, the automatic relevance
determination (ARD) \emph{squared exponential} kernel 
\begin{align}
  K_{\sf{SE}}(\brmx,\brmx') &= \theta_0\exp\left\{-\frac{1}{2}r^2(\brmx,\brmx')\right\}
  &
  r^2(\brmx,\brmx') &= \sum^D_{d=1}(x_d-x'_d)^2/\theta_d^2.
\end{align}
is often a default choice for Gaussian process regression.  However,
sample functions with this covariance function are unrealistically
smooth for practical optimization problems.  We instead propose the
use of the ARD Mat\'{e}rn~$5/2$ kernel:
\begin{align}
  K_{\sf{M52}}(\brmx,\brmx') &= \theta_0
  \left(
  1+ \sqrt{5 r^2(\brmx,\brmx')} + \frac{5}{3}r^2(\brmx,\brmx')
  \right)
  \exp\left\{-\sqrt{5 r^2(\brmx,\brmx')}\right\}.
\end{align}
This covariance function results in sample functions which are
twice differentiable, an assumption that corresponds to those made by,
e.g., quasi-Newton methods, but without requiring the smoothness of
the squared exponential.

After choosing the form of the covariance, we must also manage the
hyperparameters that govern its behavior (Note that these
``hyperparameters'' are different than the ones which are being
subjected to the overall Bayesian optimization.), as well as that of
the mean function.  For our problems of interest, typically we
would have~${D+3}$ Gaussian process hyperparameters:~$D$ length
scales~$\theta_{1:D}$, the covariance amplitude~$\theta_0$, the
observation noise~$\nu$, and a constant mean~$m$.  The most commonly
advocated approach is to use a point estimate of these parameters by
optimizing the marginal likelihood under the Gaussian
process
\begin{align*}
p(\brmy\given\{\brmx_n\}^N_{n=1},\theta,\nu,m) &=
  \distNorm(\brmy\given m\boldsymbol{1}, \bSigma_\theta + \nu\bI),
\end{align*}
where~${\brmy=[y_1, y_2, \cdots, y_n]^{\trans}}$, and~$\bSigma_\theta$
is the covariance matrix resulting from the~$N$ input points under the
hyperparameters~$\theta$.

\begin{figure}[t]
  \centering
  \begin{minipage}[t]{0.49\textwidth}
  \centering
  \subfloat[{\scriptsize Posterior samples under varying hyperparameters}]{%
    \includegraphics[width=1\textwidth]{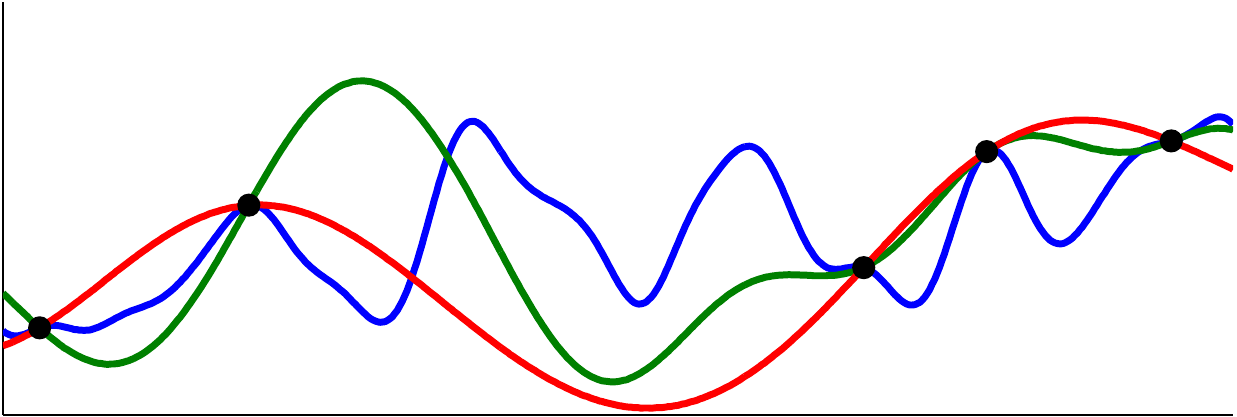}%
  }\\
  \subfloat[{\scriptsize Expected improvement under varying hyperparameters}]{%
    \includegraphics[width=1\textwidth]{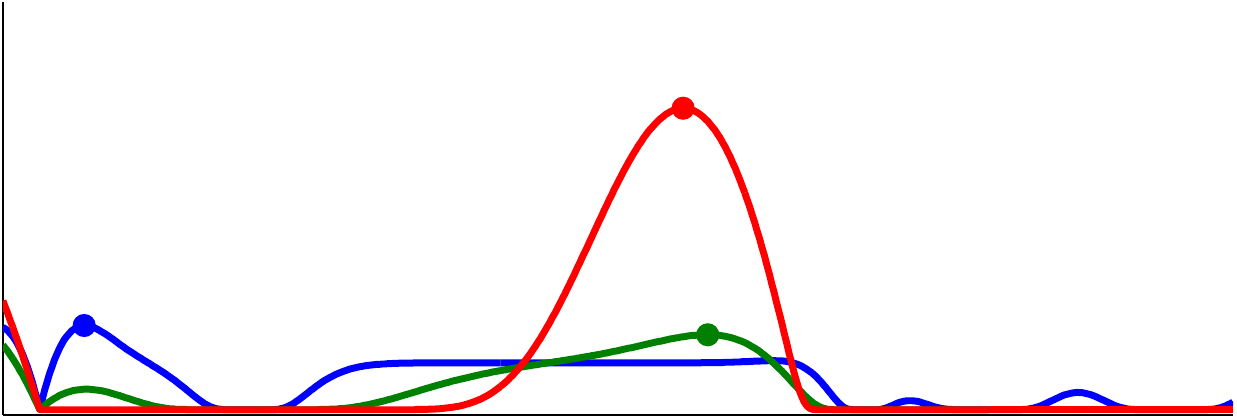}%
  }\\
  \subfloat[{\scriptsize Integrated expected improvement}]{%
    \includegraphics[width=1\textwidth]{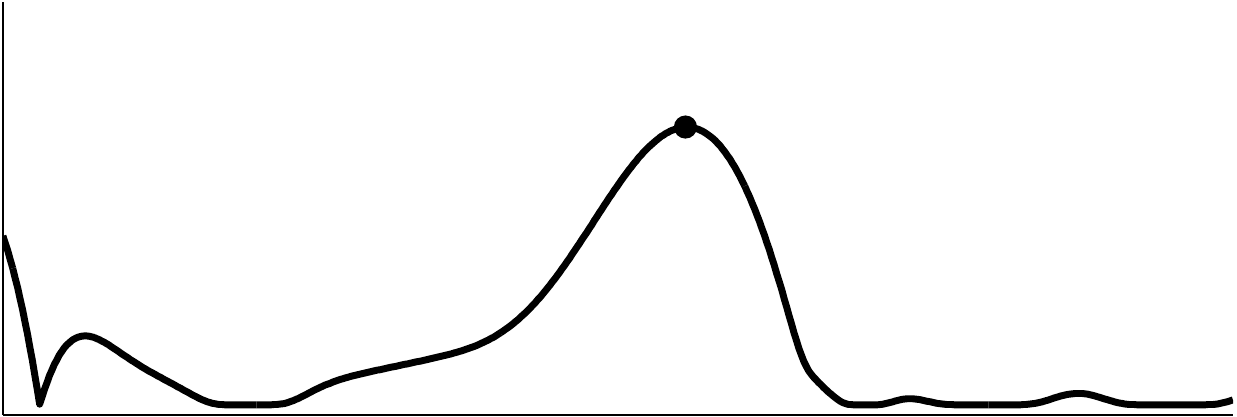}%
  }
  \caption{{\small Illustration of integrated expected improvement. (a)~Three
    posterior samples are shown, each with different length scales,
    after the same five observations.  (b)~Three expected improvement
    acquisition functions, with the same data and hyperparameters. The
    maximum of each is shown. (c)~The integrated expected improvement,
    with its maximum shown.}}
  \label{fig:illust-hyper}
  \end{minipage}~\quad
  \begin{minipage}[t]{0.49\textwidth}
  \centering
  \setcounter{subfigure}{0}
  \subfloat[{\scriptsize Posterior samples after three data}]{%
    \includegraphics[width=1\textwidth]{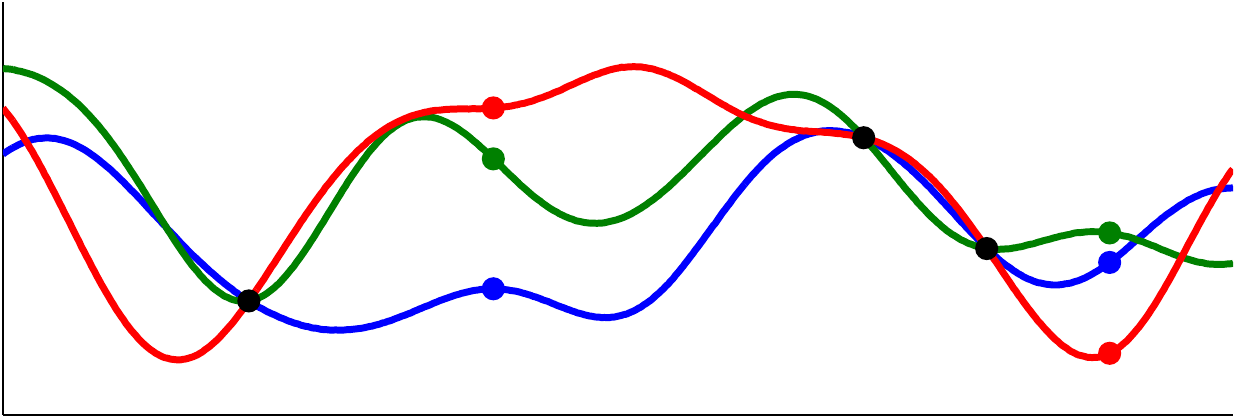}%
  }\\
  \subfloat[{\scriptsize Expected improvement under three fantasies}]{%
    \includegraphics[width=1\textwidth]{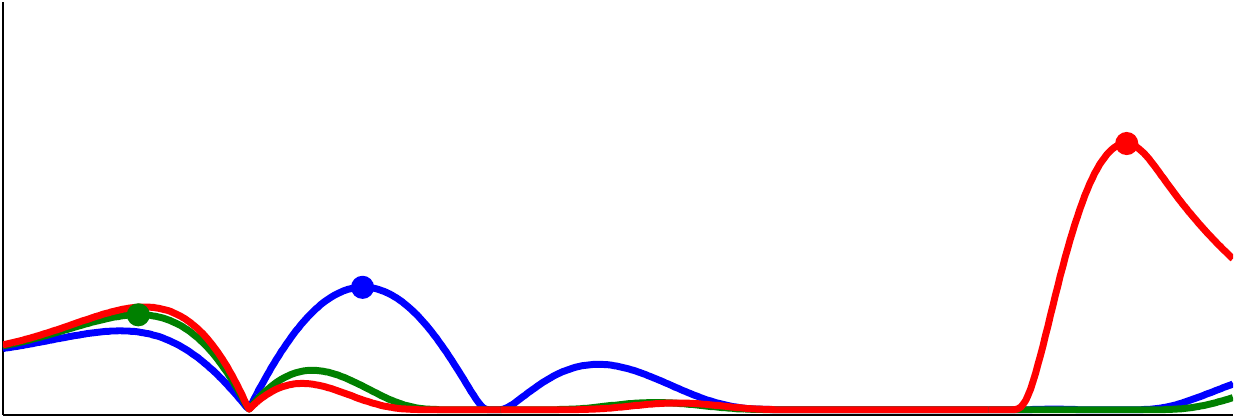}%
  }\\
  \subfloat[{\scriptsize Expected improvement across fantasies}]{%
    \includegraphics[width=1\textwidth]{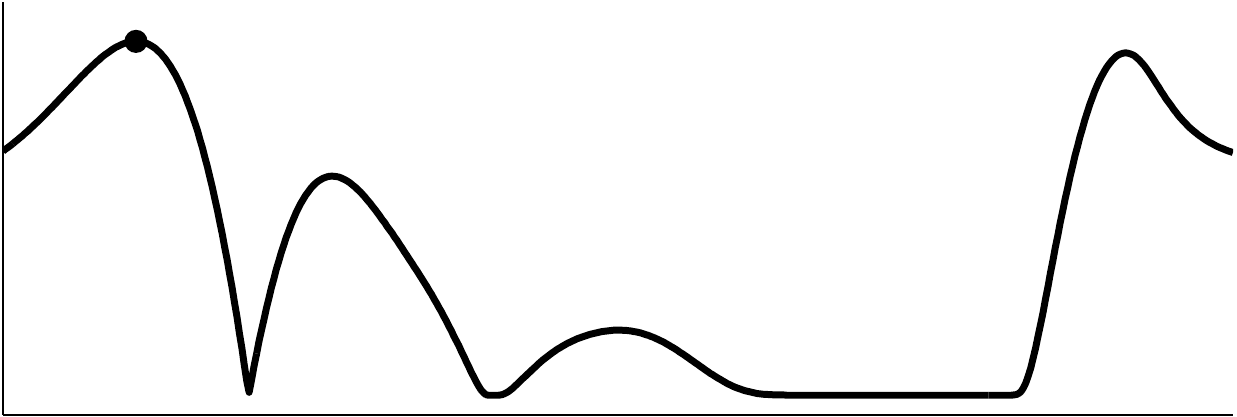}%
  }
  \caption{{\small Illustration of the acquisition with pending
    evaluations. (a)~Three data have been observed and three posterior
    functions are shown, with ``fantasies'' for three pending
    evaluations.  (b)~Expected improvement, conditioned on the each
    joint fantasy of the pending outcome.  (c)~Expected improvement
    after integrating over the fantasy outcomes.}}
  \label{fig:illust-parallel}
  \end{minipage}
\end{figure}

However, for a fully-Bayesian treatment of hyperparameters (summarized
here by~$\theta$ alone), it is desirable to marginalize over
hyperparameters and compute the \emph{integrated acquisition
  function}:

\begin{align}
  \hat{a}(\brmx\,;\,\{\brmx_n,y_n\}) &= \int a(\brmx\aqhist)\,p(\theta\given\{\brmx_n,y_n\}^N_{n=1})\,
  \mathrm{d}\theta,
\end{align}
where~$a(\brmx)$ depends on~$\theta$ and all of the observations.  For
probability of improvement and expected improvement, this expectation
is the correct generalization to account for uncertainty in
hyperparameters.  We can therefore blend acquisition functions arising
from samples from the posterior over GP hyperparameters and have a
Monte Carlo estimate of the integrated expected improvement.  These
samples can be acquired efficiently using slice sampling, as described
in~\citet{Murray-Adams-2010a}.  As both optimization
and Markov chain Monte Carlo are computationally dominated by the
cubic cost of solving an~$N$-dimensional linear system (and our
function evaluations are assumed to be much more expensive anyway),
the fully-Bayesian treatment is sensible and our empirical evaluations
bear this out.  Figure~\ref{fig:illust-hyper} shows how the integrated
expected improvement changes the acquistion function.

\subsection{Modeling Costs}
Ultimately, the objective of Bayesian optimization is to find a good
setting of our hyperparameters as quickly as possible.  Greedy
acquisition procedures such as expected improvement try to make the
best progress possible in the next function evaluation.  From a
practial point of view, however, we are not so concerned with function
evaluations as with wallclock time.  Different regions of the
parameter space may result in vastly different execution times, due to
varying regularization, learning rates, etc.  To improve our
performance in terms of wallclock time, we propose optimizing with the
\emph{expected improvement per second}, which prefers to acquire
points that are not only likely to be good, but that are also likely
to be evaluated quickly.  This notion of cost can be naturally
generalized to other budgeted resources, such as reagents or money.

Just as we do not know the true objective function~$f(\brmx)$, we also
do not know the \emph{duration
  function}~${c(\brmx):\mcX\to\reals^{+}}$.  We can nevertheless
employ our Gaussian process machinery to model~$\ln c(\brmx)$
alongside~$f(\brmx)$.  In this work, we assume that these functions
are independent of each other, although their coupling may be usefully
captured using GP variants of multi-task learning (e.g., \citet{teh-etal-2005a,bonilla-etal-2008a}).
Under the independence assumption, we can easily compute the predicted
expected inverse duration and use it to compute the expected
improvement per second as a function of~$\brmx$.

\subsection{Monte Carlo Acquisition for Parallelizing Bayesian Optimization}
With the advent of multi-core computing, it is natural to ask how we
can parallelize our Bayesian optimization procedures.  More generally
than simply batch parallelism, however, we would like to be able to
decide what~$\brmx$ should be evaluated next, even while a set of
points are being evaluated.  Clearly, we cannot use the same
acquisition function again, or we will repeat one of the pending
experiments.  We would ideally perform a roll-out of our acquisition
policy, to choose a point that appropriately balanced information gain
and exploitation.  However, such roll-outs are generally intractable.
Instead we propose a sequential strategy that takes advantage of the
tractable inference properties of the Gaussian process to compute
Monte Carlo estimates of the acquisiton function under different
possible results from pending function evaluations.

Consider the situation in which~$N$ evaluations have completed,
yielding data~$\{\brmx_n,y_n\}^N_{n=1}$, and in which~$J$ evaluations
are pending at locations~$\{\brmx_j\}^J_{j=1}$.  Ideally, we would
choose a new point based on the expected acquisition function under
all possible outcomes of these pending evaluations:
\begin{multline}
  \hat{a}(\brmx\aqhist,\{\brmx_j\}) = \\\int_{\reals^J}
  a(\brmx\aqhist,\{\brmx_j,y_j\})\,p(\{y_j\}^J_{j=1}
  \given\{\brmx_j\}^J_{j=1},\{\brmx_n,y_n\}^N_{n=1})
  \,
  \mathrm{d}y_1\cdots\mathrm{d}y_J.
\end{multline}
This is simply the expectation of~$a(\brmx)$ under a~$J$-dimensional
Gaussian distribution, whose mean and covariance can easily be
computed.  As in the covariance hyperparameter case, it is
straightforward to use samples from this distribution to compute the
expected acquisition and use this to select the next point.
Figure~\ref{fig:illust-parallel} shows how this procedure would
operate with queued evaluations.  We note that a similar approach is
touched upon briefly by \citet{Ginsbourger2010a}, but they view it as
too intractable to warrant attention.  We have found our Monte Carlo
estimation procedure to be highly effective in practice, however, as
will be discussed in Section~\ref{sec:empirical}.

\section{Empirical Analyses}
In this section, we empirically analyse\footnote{All experiments were
  conducted on identical machines using the Amazon EC2 service.} the
algorithms introduced in this paper and compare to existing strategies
and human performance on a number of challenging machine learning
problems.  We refer to our method of expected improvement while
marginalizing GP hyperparameters as ``GP EI MCMC'', optimizing
hyperparameters as ``GP EI Opt'', EI per second as ``GP EI per
Second'', and $N$ times parallelized GP EI MCMC as ``$N$x GP EI
MCMC''.

\label{sec:empirical}

\subsection{Branin-Hoo and Logistic Regression}
\label{sec:branin}
\begin{figure}[t]
\begin{center}
\subfloat[\label{fig:braninhoo_results}]{
\includegraphics[width=0.5\textwidth]{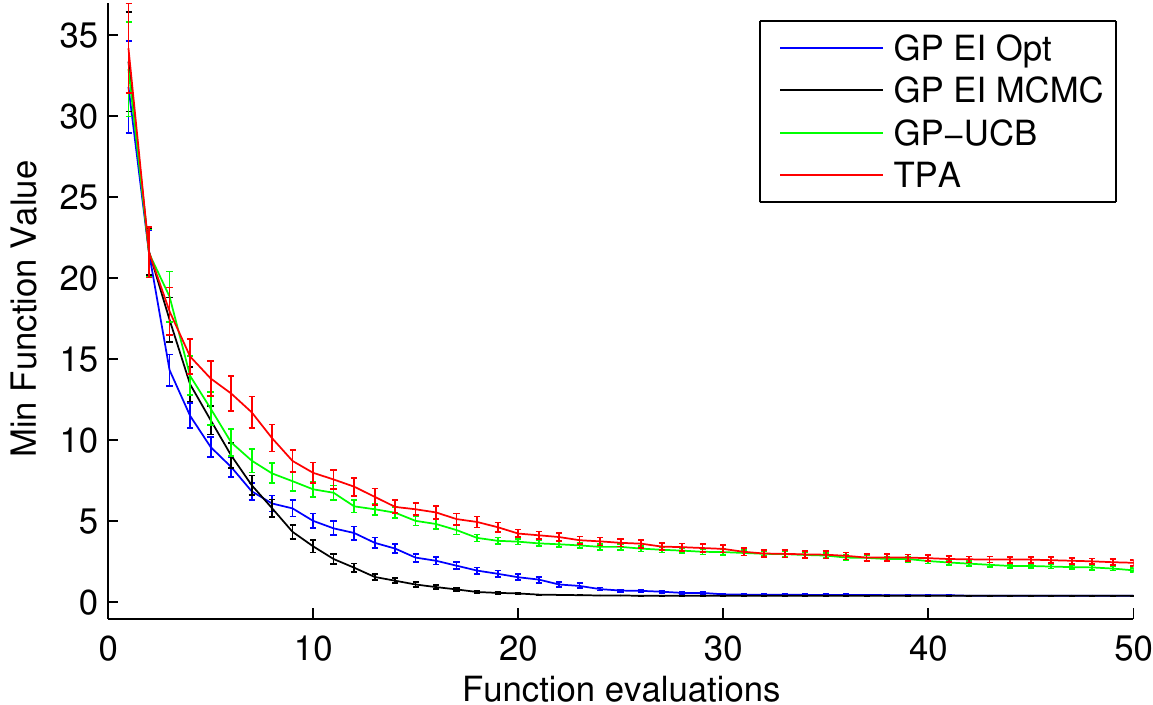}}
\subfloat[\label{fig:logreg_results}]{
\includegraphics[width=0.5\textwidth]{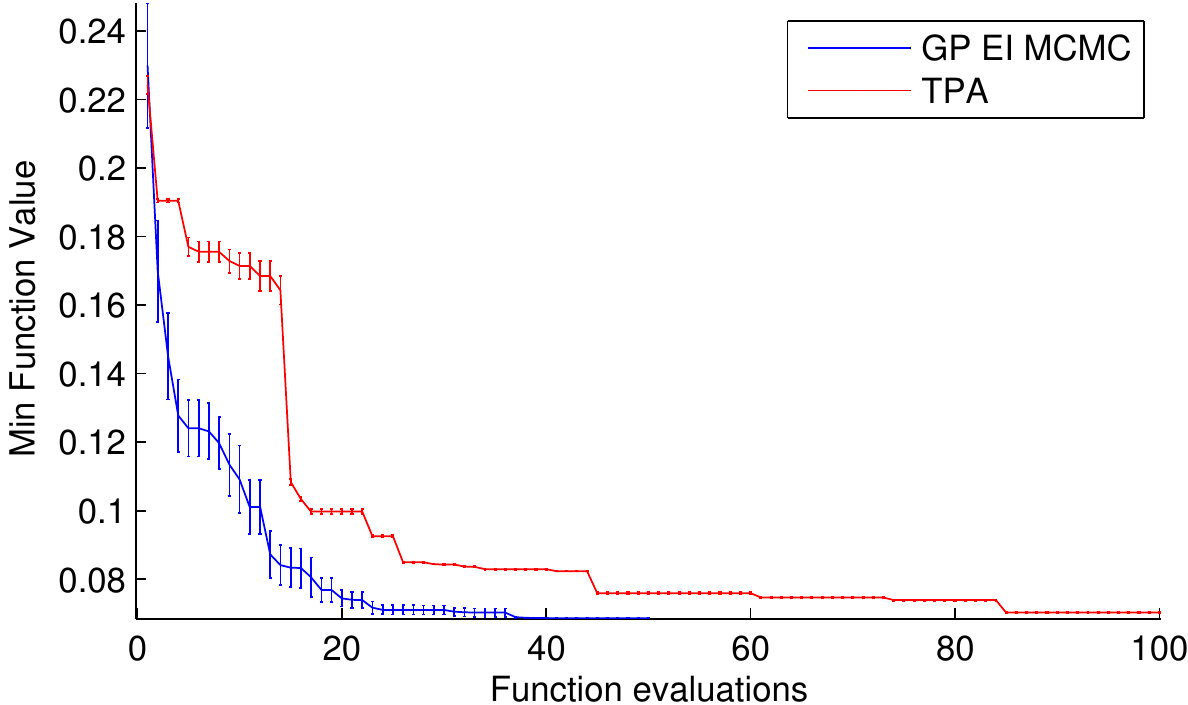}}

\end{center}
\caption{A comparison of standard approaches compared to our GP EI MCMC approach on the Branin-Hoo function (\ref{fig:braninhoo_results}) and training logistic regression on the MNIST data (\ref{fig:logreg_results}).}
\end{figure}

We first compare to standard approaches and the recent Tree Parzen
Algorithm\footnote{Using the publicly available code from
  \url{https://github.com/jaberg/hyperopt/wiki}} (TPA) of
\citet{BergstraJ2011} on two standard problems.  The Branin-Hoo
function is a common benchmark for Bayesian optimization
techniques~\citep{Jones2001} that is defined over $x \in \reals^2$
where $0 \leq x_1 \leq 15$ and $-5 \leq x_2 \leq 15$.  We also compare
to TPA on a logistic regression classification task on the popular
MNIST data. The algorithm requires choosing four hyperparameters, the
learning rate for stochastic gradient descent, on a log scale from 0
to 1, the~$\ell_2$ regularization parameter, between 0 and 1, the mini batch
size, from 20 to 2000 and the number of learning epochs, from 5 to
2000.  Each algorithm was run on the Branin-Hoo and logistic
regression problems 100 and 10 times respectively and mean and
standard error are reported.  The results of these analyses are
presented in Figures \ref{fig:braninhoo_results} and
\ref{fig:logreg_results} in terms of the number of times the function
is evaluated.  On Branin-Hoo, integrating over hyperparameters
is superior to using a point estimate and the GP EI significantly
outperforms TPA, finding the minimum in fewer than half as many
evaluations, in both cases.

\subsection{Online LDA}
\begin{figure}[ht]
\begin{center}
\subfloat[\label{fig:lda_fun_evals}]{
\includegraphics[width=0.5\textwidth]{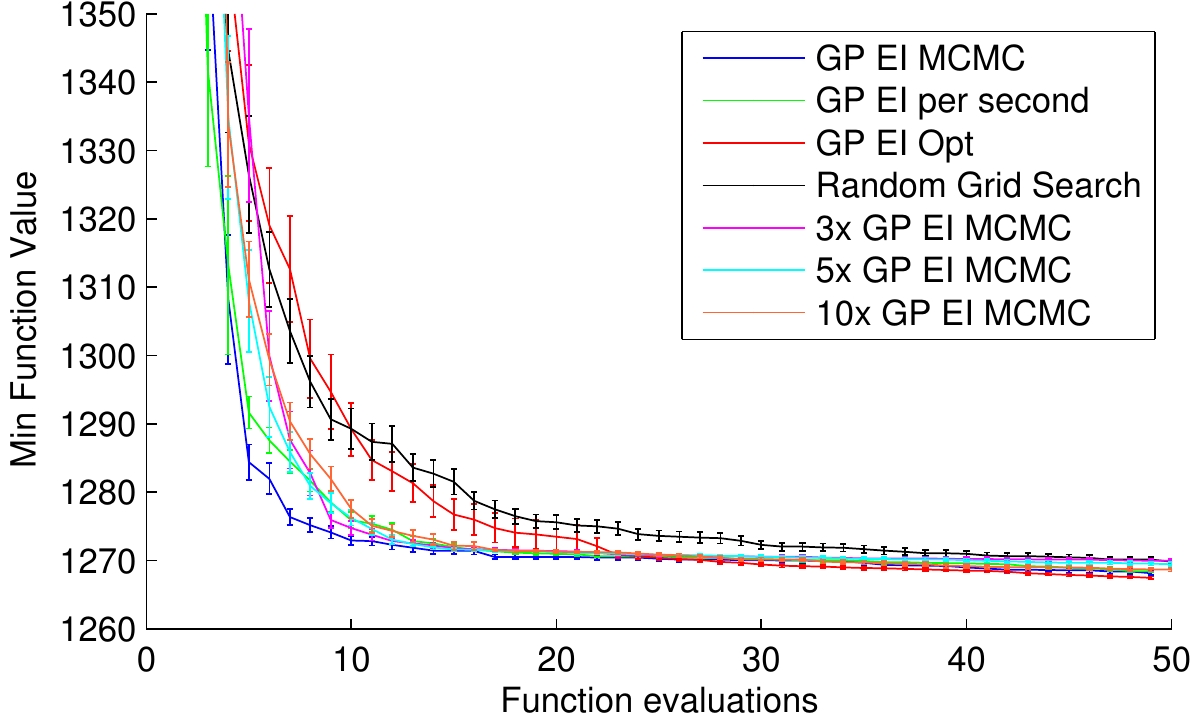}}
\subfloat[\label{fig:lda_seconds}]{
\includegraphics[width=0.5\textwidth]{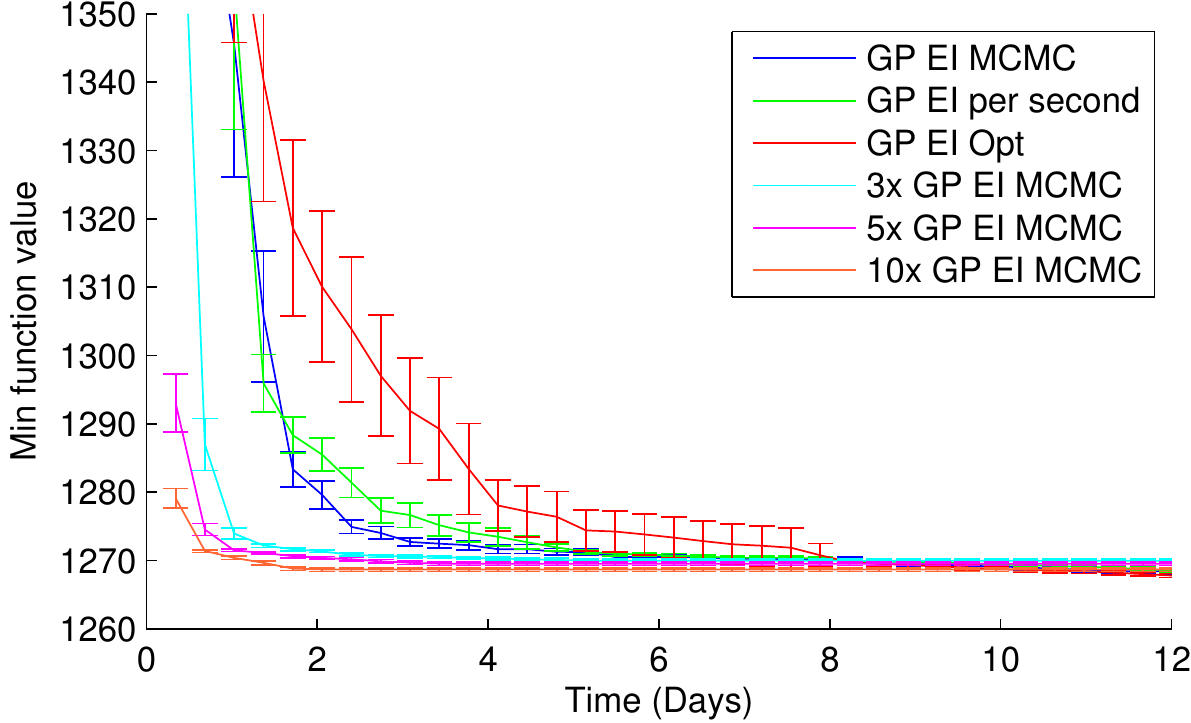}}\\
\subfloat[\label{fig:lda_gridvsnogrid}]{
\includegraphics[width=0.5\textwidth]{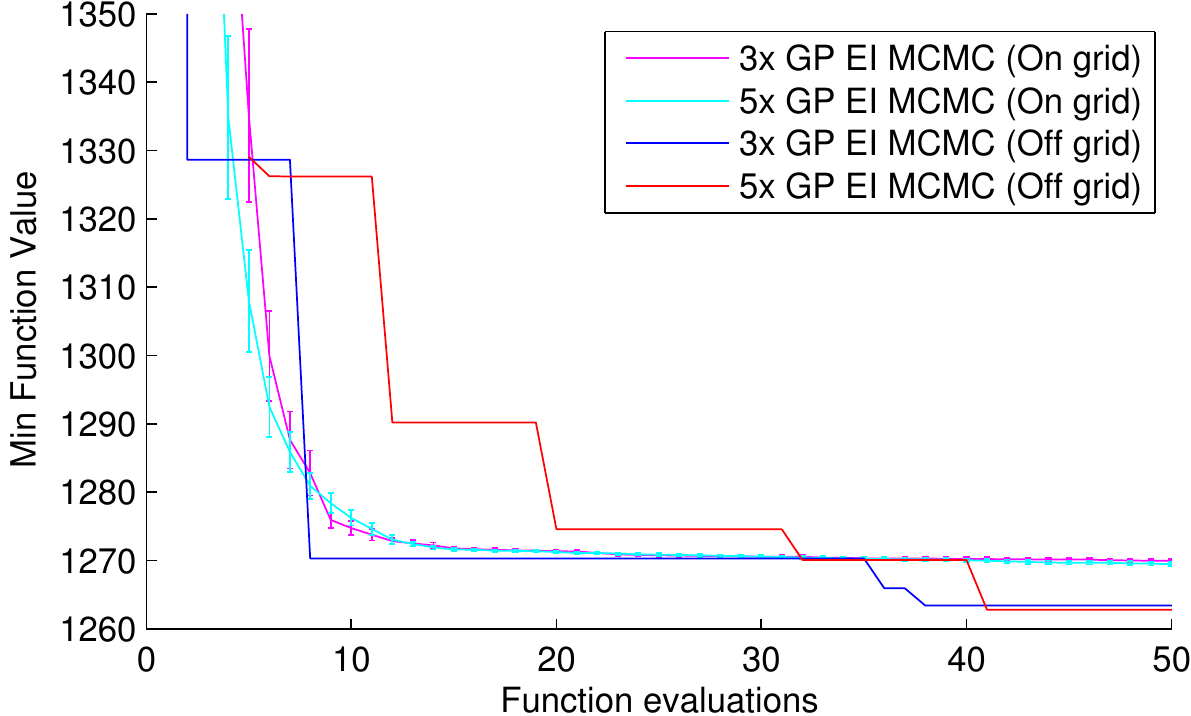}}
\end{center}
\caption{Different strategies of optimization on the Online LDA
  problem compared in terms of function evaluations
  (\ref{fig:lda_fun_evals}), walltime (\ref{fig:lda_seconds}) and
  constrained to a grid or not (\ref{fig:lda_gridvsnogrid}).}
\end{figure}

Latent Dirichlet allocation (LDA) is a directed graphical model for
documents in which words are generated from a mixture of multinomial
``topic'' distributions. Variational Bayes is a popular paradigm for
learning and, recently, \citet{Hoffman2010} proposed an online
learning approach in that context. Online LDA requires two learning
parameters,~$\tau_0$ and~$\kappa$, that control the learning
rate~${\rho_t = (\tau_0 + t)^{-\kappa}}$ used to update the
variational parameters of LDA based on the $t^{\rm th}$ minibatch of
document word count vectors. The size of the minibatch is also a third
parameter that must be chosen. \citet{Hoffman2010} relied on an
exhaustive grid search of size~${6\times6\times8}$, for a total of 288
hyperparameter configurations.

We used the code made publically available by \citet{Hoffman2010} to
run experiments with online LDA on a collection of Wikipedia
articles. We downloaded a random set of 249,560 articles, split into
training, validation and test sets of size 200,000, 24,560 and 25,000
respectively. The documents are represented as vectors of word counts
from a vocabulary of 7,702 words. As reported in \citet{Hoffman2010},
we used a lower bound on the per word perplixity of the validation set
documents as the performance measure. One must also specify the number
of topics and the hyperparameters $\eta$ for the symmetric Dirichlet
prior over the topic distributions and $\alpha$ for the symmetric
Dirichlet prior over the per document topic mixing weights. We
followed \citet{Hoffman2010} and used 100 topics and~${\eta = \alpha =
0.01}$ in our experiments in order to emulate their analysis and
repeated exactly the grid search reported in the
paper\footnote{i.e. the only difference was the randomly sampled
  collection of articles in the data set and the choice of the
  vocabulary. We ran each evaluation for 10 hours or until
  convergence.}. Each online LDA evaluation generally took between
five to ten hours to converge, thus the grid search requires
approximately 60 to 120 processor days to complete.

In Figures~\ref{fig:lda_fun_evals} and \ref{fig:lda_seconds} we
compare our various strategies of optimization over the same grid on
this expensive problem. That is, the algorithms were restricted to
only the exact parameter settings as evaluated by the grid search.
Each optimization was then repeated one hundred times (each time
picking two different random experiments to initialize the
optimization with) and the mean and standard error are
reported. Figures~\ref{fig:lda_fun_evals} and \ref{fig:lda_seconds}
respectively show the average minimum loss (perplexity) achieved by
each strategy compared to the number of times online LDA is evaluated
with new parameter settings and the duration of the optimization in
days.  Figure~\ref{fig:lda_gridvsnogrid} shows the average loss of 3
and 5 times parallelized GP EI MCMC which are restricted to the same
grid as compared to a single run of the same algorithms where the
algorithm can flexibly choose new parameter settings within the same
range by optimizing the expected improvement.

In this case, integrating over hyperparameters is superior to using a point
estimate.  While GP EI MCMC is the most efficient in
terms of function evaluations, we see that parallelized GP EI MCMC
finds the best parameters in significantly less time.  Finally, in
Figure~\ref{fig:lda_gridvsnogrid} we see that the parallelized GP EI
MCMC algorithms find a significantly better minimum value than was
found in the grid search used by \citet{Hoffman2010} while running
a fraction of the number of experiments.

\subsection{Motif Finding with Structured Support Vector Machines}
\begin{figure}[ht]
\begin{center}
\subfloat[\label{fig:motif_seconds}]{
  \includegraphics[width=0.5\textwidth]{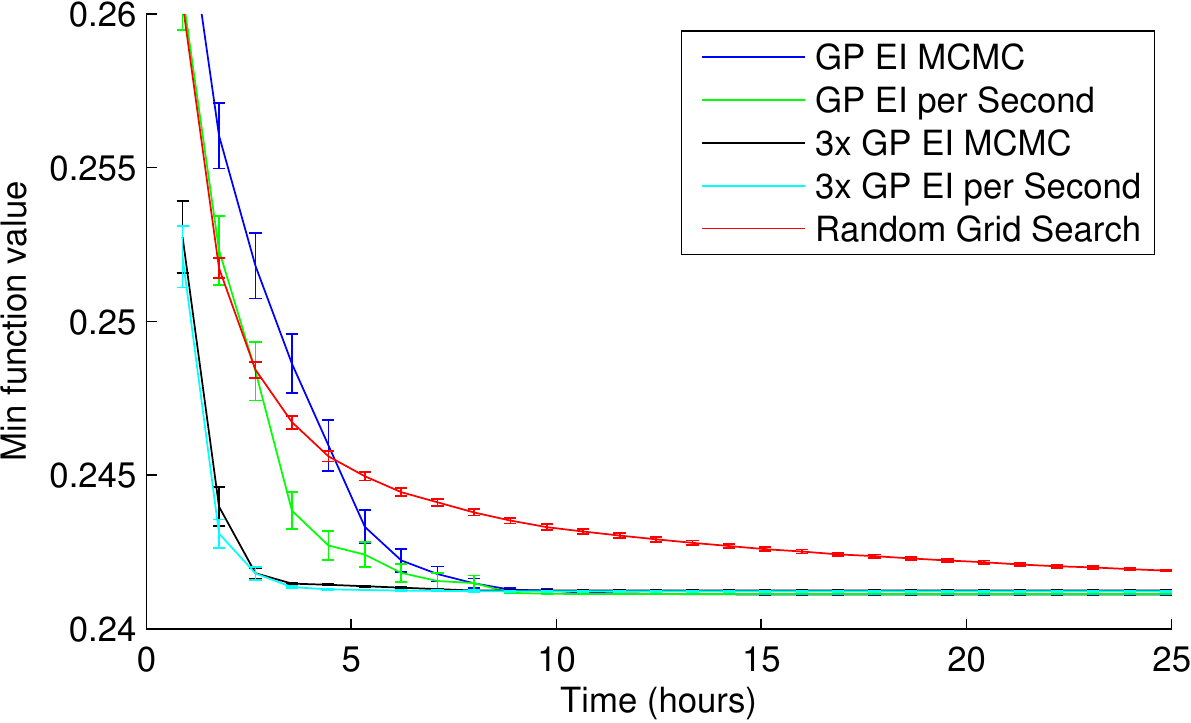}}
\subfloat[\label{fig:motif_evals}]{
  \includegraphics[width=0.5\textwidth]{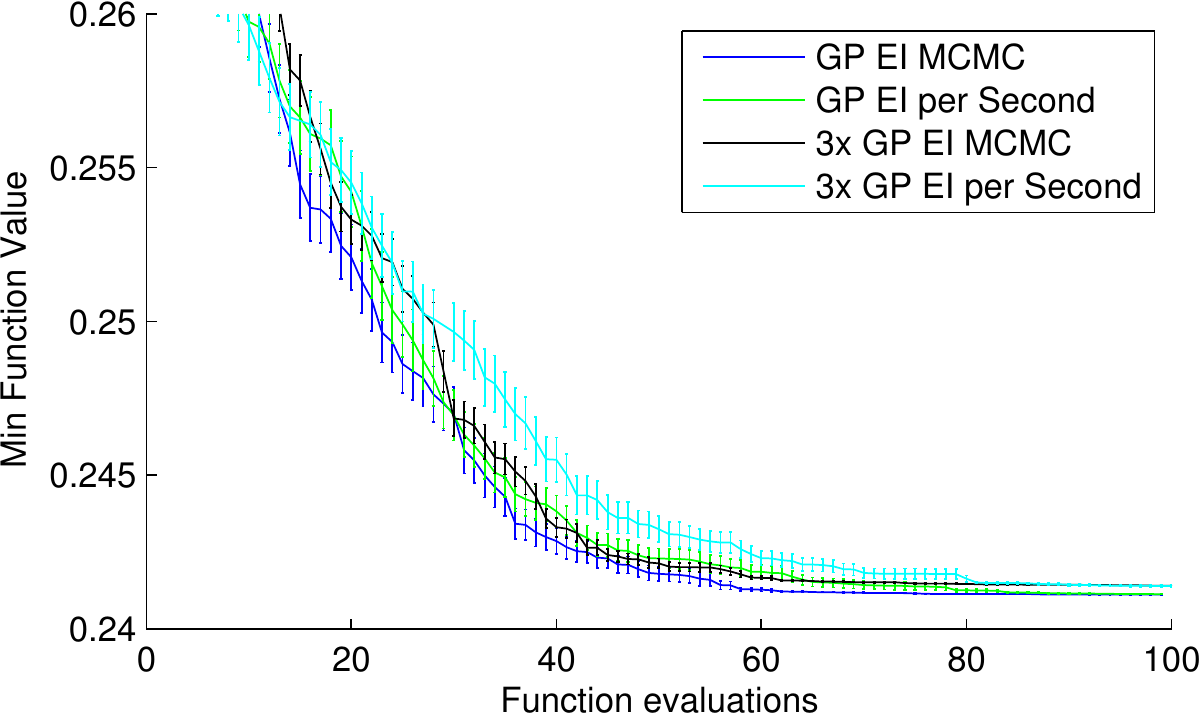}}\\
\subfloat[\label{fig:motif_kernels}]{
  \includegraphics[width=0.5\textwidth]{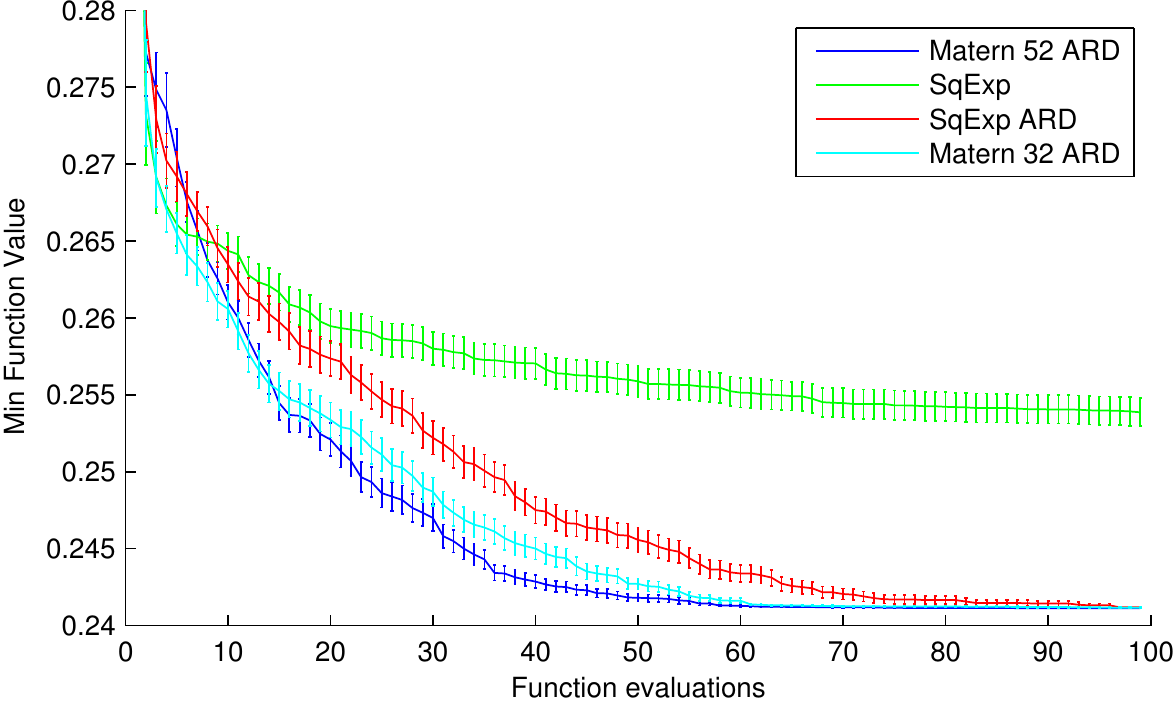}}
\end{center}
\caption{A comparison of various strategies for optimizing the
  hyperparameters of M3E models on the protein motif finding task in
  terms of wallclock time (\ref{fig:motif_seconds}), function evaluations
  (\ref{fig:motif_evals}) and different covariance functions(\ref{fig:motif_kernels}).}
\end{figure}

In this example, we consider optimizing the learning parameters of
Max-Margin Min-Entropy (M3E) Models~\citep{Miller-etal-2012}, which
include Latent Structured Support Vector
Machines~\citep{Yu-Joachims-2009} as a special case.  Latent structured
SVMs outperform SVMs on problems where they can explicitly model
problem-dependent hidden variables. A popular example task is the
binary classification of protein DNA
sequences~\citep{Miller-etal-2012,Kumar-etal-2010,Yu-Joachims-2009}.
The hidden variable to be modeled is the unknown location of
particular subsequences, or \emph{motifs}, that are indicators of
positive sequences.

Setting the hyperparameters, such as the regularisation term, $C$, of
structured SVMs remains a challenge and these are typically set
through a time consuming grid search procedure as is done in
\citet{Miller-etal-2012} and \citet{Yu-Joachims-2009}. Indeed,
\citet{Kumar-etal-2010} report that hyperparameter selection was
avoided for the motif finding task due to being too computationally
expensive.  However, \citet{Miller-etal-2012} demonstrate that
classification results depend highly on the setting of the parameters,
which differ for each protein.

M3E models introduce an entropy term, parameterized by $\alpha$, which
enables the model to significantly outperform latent structured
SVMs. This additional performance, however, comes at the expense of an
additional problem-dependent hyperparameter.  We emulate the
experiments of \citet{Miller-etal-2012} for one protein with
approximately 40,000 sequences. We explore 25 settings of the
parameter $C$, on a log scale from $10^{-1}$ to $10^6$, 14 settings of
$\alpha$, on a log scale from 0.1 to 5 and the model convergence
tolerance, ${\epsilon \in
\{10^{-4},\,10^{-3},\,10^{-2},\,10^{-1}\}}$.  We ran a grid search
over the 1,400 possible combinations of these parameters, evaluating
each over 5 random 50-50 training and test splits.

In Figures~\ref{fig:motif_seconds} and \ref{fig:motif_evals}, we
compare the randomized grid search to GP EI MCMC, GP EI per Second and
their 3x parallelized versions, all constrained to the same points on
the grid, in terms of minimum validation error vs wallclock time and
function evaluations. Each algorithm was repeated 100 times and the
mean and standard error are shown. We observe that the Bayesian
optimization strategies are considerably more efficient than grid
search which is the status quo. In this case, GP EI MCMC is superior
to GP EI per Second in terms of function evaluations but GP EI per
Second finds better parameters faster than GP EI MCMC as it learns to
use a less strict convergence tolerance early on while exploring the
other parameters. Indeed, 3x GP EI per second is the least efficient
in terms of function evaluations but finds better parameters faster
than all the other algorithms.

Figure \ref{fig:motif_kernels} compares the use of various covariance
functions in GP EI MCMC optimization on this problem.  The
optimization was repeated for each covariance 100 times and the mean
and standard error are shown.  It is clear that the selection of an
appropriate covariance significantly affects performance and the
estimation of length scale parameters is critical.  The assumption of
the infinite differentiability of the underlying function as imposed
by the commonly used squared exponential is too restrictive for this
problem.

\subsection{Convolutional Networks on CIFAR-10}
\begin{figure}[ht]
\begin{center}
\subfloat[\label{fig:convnet_fun_evals}]{
\includegraphics[width=0.5\textwidth]{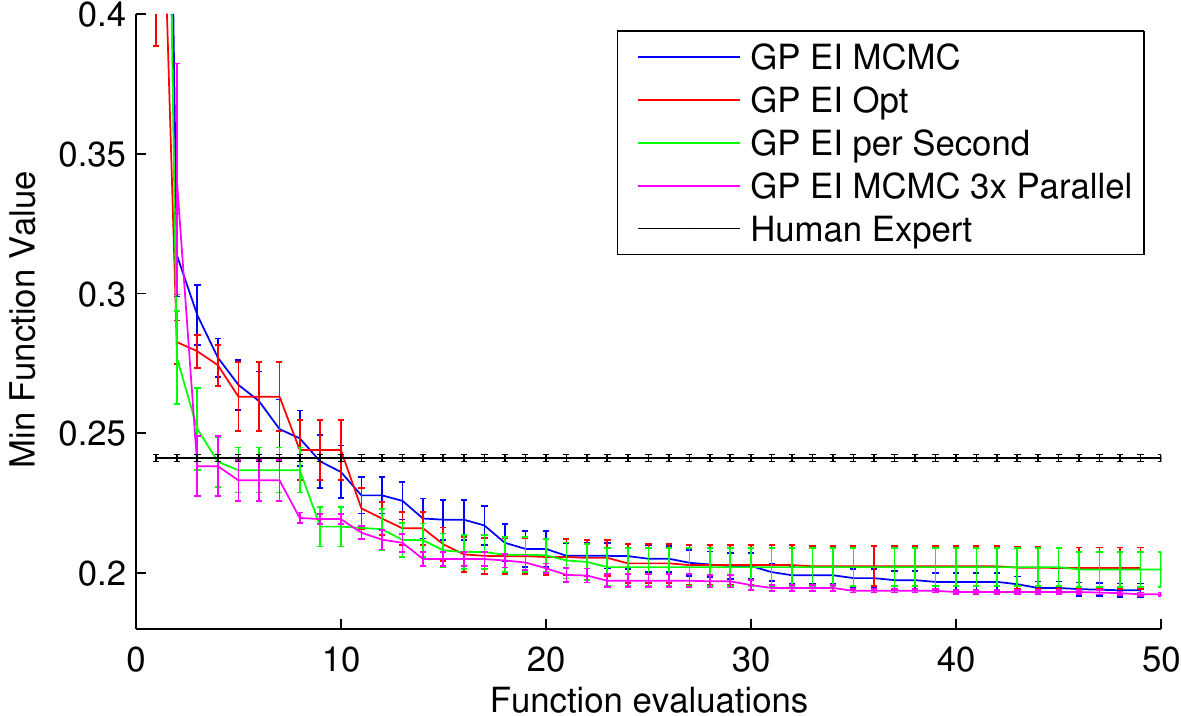}}
\subfloat[\label{fig:convnet_seconds}]{
\includegraphics[width=0.5\textwidth]{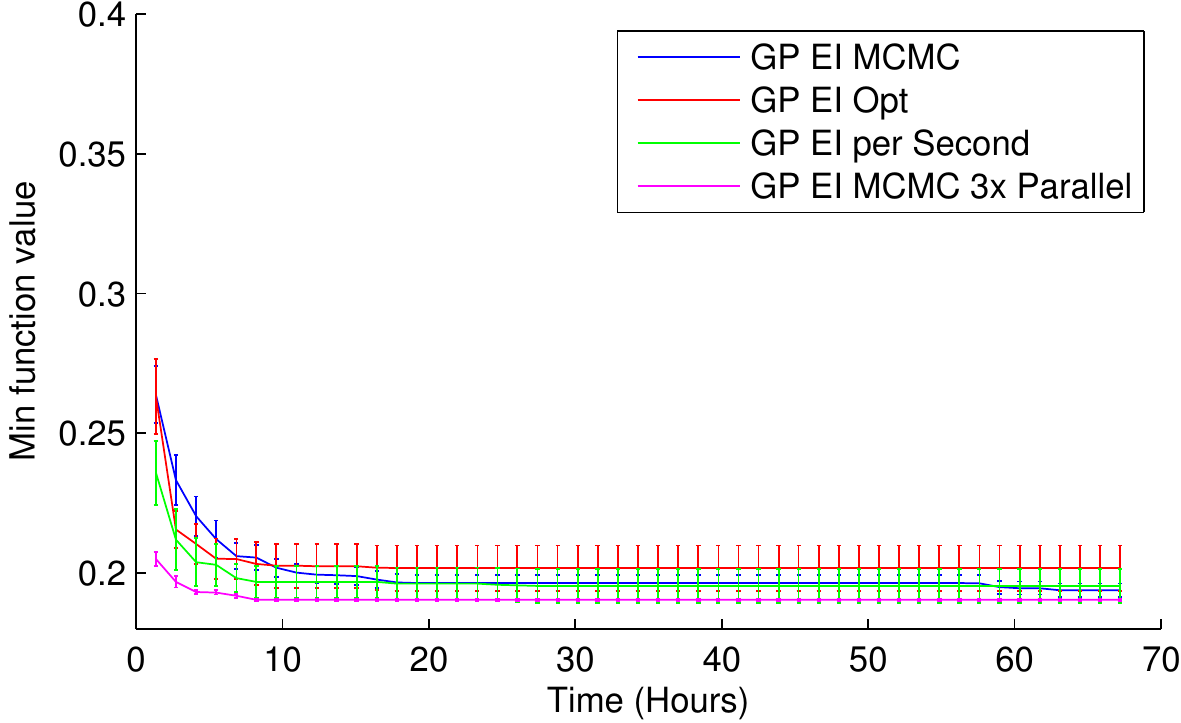}}
\end{center}
\caption{Validation error on the CIFAR-10 data for different optimization strategies.}
\label{fig:convnet}
\end{figure}

Neural networks and deep learning methods notoriously require careful
tuning of numerous hyperparameters.  Multi-layer convolutional neural
networks are an example of such a model for which a thorough
exploration of architechtures and hyperparameters is beneficial, as
demonstrated in~\citet{Saxe-etal-2011}, but often computationally
prohibitive.  While \citet{Saxe-etal-2011} demonstrate a methodology
for efficiently exploring model architechtures, numerous
hyperparameters, such as regularisation parameters, remain. In this
empirical analysis, we tune nine hyperparameters of a three-layer
convolutional network, described in~\citet{Krizhevsky-2009a} on the
CIFAR-10 benchmark dataset using the code provided\footnote{Available
  at: \url{http://code.google.com/p/cuda-convnet/} using the architechture
  defined in
  \url{http://code.google.com/p/cuda-convnet/source/browse/trunk/example-layers/layers-18pct.cfg}}.
This model has been carefully tuned by a human
expert~\citep{Krizhevsky-2009a} to achieve a highly competitive result
of 18\% test error, which matches the published state of the
art\footnote{Without augmenting the training data.}
result~\citep{Coates-2011a} on CIFAR-10. The parameters we explore
include the number of epochs to run the model, the learning rate, four
weight costs (one for each layer and the softmax output weights), and
the width, scale and power of the response normalization on the
pooling layers of the network.

We optimize over the nine parameters for each strategy on a withheld
validation set and report the mean validation error and standard error
over five separate randomly initialized runs.  Results are presented
in Figure \ref{fig:convnet} and contrasted with the average results
achieved using the best parameters found by the expert.  The best
hyperparameters\footnote{The optimized parameters deviate
  interestingly from the expert-determined settings; 
  e.g., the optimal weight costs are asymmetric (the weight cost of the
  second layer is approximately an order of magnitude smaller than the
  first layer), a learning rate two orders of magnitude smaller,
  a slightly wider response normalization, larger scale and much
  smaller power.} found by the GP EI MCMC approach achieve an error on
the \emph{test set} of~$14.98\%$, which is over 3\% better than the
expert and the state of the art on CIFAR-10.

\section{Conclusion}
In this paper we presented methods for performing Bayesian
optimization of hyperparameters associated with general machine
learning algorithms.  We introduced a fully Bayesian treatment for
expected improvement, and algorithms for dealing with variable time
regimes and parallelized experiments.  Our empirical analysis
demonstrates the effectiveness of our approaches on three challenging
recently published problems spanning different areas of machine
learning.  The code used will be made publicly available.  The
resulting Bayesian optimization finds better hyperparameters
significantly faster than the approaches used by the authors. Indeed
our algorithms \emph{surpassed} a human expert at selecting
hyperparameters on the competitive CIFAR-10 dataset and as a result
beat the state of the art by over 3\%.

\subsection*{Acknowledgements}
This work was supported by a grant from Amazon Web Services and by a
DARPA Young Faculty Award.

\bibliographystyle{unsrtnat} \bibliography{draft}

\end{document}